\documentclass{article}

 \usepackage[preprint]{neurips_2026}

\usepackage[utf8]{inputenc} 
\usepackage[T1]{fontenc}    
\usepackage{hyperref}       
\usepackage{url}            
\usepackage{booktabs}       
\usepackage{amsfonts}       
\usepackage{nicefrac}       
\usepackage{microtype}      
\usepackage{xcolor}         
\usepackage{graphicx}
\usepackage{bbm}
\usepackage{multirow}
\usepackage{bm}
\usepackage[table]{xcolor}
\usepackage{amsmath}
\usepackage{pifont}
\usepackage{wrapfig}  
\usepackage{subcaption}
\usepackage{caption} 
\usepackage{placeins} 

\usepackage{hyperref}
\hypersetup{
    colorlinks=true,
    urlcolor=blue,
    citecolor=blue,
    linkcolor=red
}
\title{Preserving DEG Rankings for Gene Discovery \\ in Histology-Based Spatial Gene Expression Prediction}

\author{
Kaito Shiku$^{1}$ \quad
Kazuya Nishimura$^{2}$ \quad
Yasuhiro Kojima$^{3}$ \quad
Ryoma Bise$^{1}$ \quad
\\
$^{1}$ Kyushu University, Japan \quad
$^{2}$ The University of Osaka, Japan  \quad
$^{3}$ National Cancer Center, Japan \\
\texttt{kaito.shiku@human.ait.kyushu-u.ac.jp} 
}

\begin{document}

\maketitle

\begin{abstract} 
Predicting spatial gene expression from histology images could scale spatial transcriptomics (ST) to image-only cohorts, but conventional histology-based ST prediction is trained and evaluated mainly by per-gene spatial-profile reconstruction. This objective is misaligned with a key downstream use of ST: differentially expressed gene (DEG) discovery, where genes are ranked for a biological or morphology-defined contrast by evidence of between-group expression differences. We formulate image-based differential expression ranking (IDER), which asks whether predicted expression profiles preserve the contrast-specific ranked gene list obtained from measured profiles. IDER compares gene rankings induced by differential-expression statistics, rather than raw expression magnitudes or per-gene spatial correlations. We further introduce a differentiable IDER objective that aligns these statistics across genes and can be trained with morphology-derived proxy contrasts without predefined biological group labels. Experiments on public ST datasets show improved DEG-ranking agreement and pathway-enrichment overlap over conventional reconstruction objectives, including morphology-derived and pathologist-annotated tissue-region evaluations. 
The code is available at \url{https://github.com/Shiku-Kaito/IDER}.
\end{abstract}

\section{Introduction}

Predicting spatial gene expression from routine histopathology images is a promising way to extend molecular analysis to large image-only cohorts~\cite{he2020integrating,xie2023spatially,pang2021leveraging, yang2023exemplar,yang2024spatial}.
Histopathology images are widely collected in clinical and research settings, but they do not directly provide the gene expression measurements needed to interpret tissue biology. Recent methods address this gap by learning from paired histology images and spatial transcriptomics (ST) measurements, where ST provides spatially resolved gene expression while preserving tissue organization~\cite{marx2021method}. If reliable, such models could enable molecular analysis in cohorts where sequencing-based ST assays are
unavailable or costly.

\begin{figure}
    \centering
      \includegraphics[width=\linewidth]{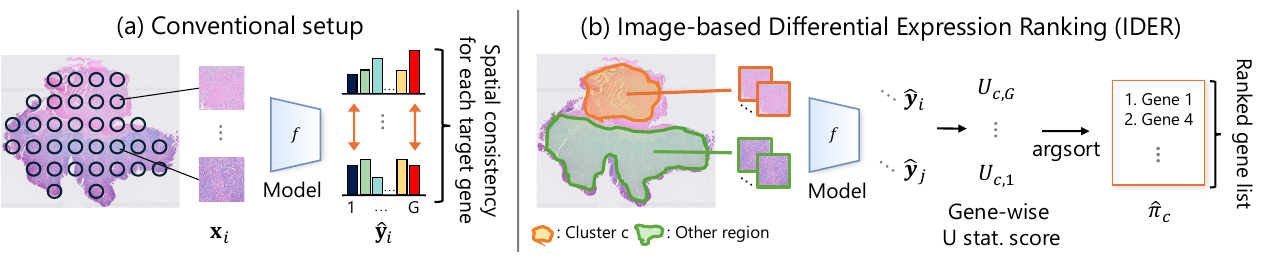}
      \caption{(a) Conventional evaluation measures spatial-profile agreement within each gene across spots. (b) Image-based Differential Expression Ranking (IDER) ranks genes by contrast statistics for a biological or morphology-defined comparison and evaluates whether predicted profiles preserve this ranked list of DEG candidates.}
      \label{fig:intro}
\end{figure}

Conventional objectives and evaluation protocols for histology-based ST prediction primarily assess whether spatial expression profiles are accurately reconstructed. Pearson correlation coefficient (PCC)-based losses and evaluation metrics evaluate, for each gene independently, whether the predicted expression pattern across spots is consistent with the measured pattern, and then average this agreement across genes~\cite{chung2024accurate, yang2023exemplar, yang2024spatial, shiku2026auxiliary}. While this is a natural criterion for expression reconstruction, it emphasizes agreement along the spatial axis within each gene rather than identifying which genes should be prioritized for downstream biological discovery. Figure~\ref{fig:intro} (a) illustrates this conventional setup, where each target gene is evaluated separately based on spatial-profile agreement.

However, a central downstream goal of ST analysis is gene discovery: identifying genes whose expression differences are associated with biologically meaningful tissue contrasts, such as those across tissue regions, disease-associated phenotypes, or morphology-defined groups~\cite{cable2022cell, walker2022deciphering}. This perspective goes beyond expression reconstruction by asking which genes should be prioritized for biological interpretation.
In differential expression analysis~\cite{robinson2010edger,ritchie2015limma, cable2022cell}, researchers specify such biological contrasts and seek to identify, from thousands of measured genes, those with strong statistical evidence of between-group expression differences. To this end, differential expression analysis assigns each gene a group-contrast statistic and ranks genes by the strength of this evidence. The top-ranked genes are reported as candidate differentially expressed genes (DEGs) or marker genes that help explain the tissue difference, support pathway interpretation, and guide downstream biological validation. 
Thus, a key output of differential expression analysis is a contrast-specific ranked gene list.\footnote{
While standard DEG analysis often considers both up- and
down-regulated genes, our study focuses on identifying genes
that are more highly expressed in a particular tissue
morphology than in other tissue groups.
}

This creates a direct mismatch between conventional reconstruction criteria and image-based DEG discovery. Conventional criteria compare predicted and measured expression along the spatial axis within each gene, and then summarize these per-gene agreements. In differential expression analysis, however, the relevant output is defined along the gene axis: genes are ranked against one another by their contrast scores for a biological or morphology-defined comparison. The missing evaluation axis is therefore whether predicted expression profiles induce the same contrast-specific ranked gene list as measured profiles. Errors in this ordering can change which genes are reported as candidate DEGs or marker genes, and consequently, which pathways and biological hypotheses are carried forward.

To address this limitation, we formulate image-based differential expression ranking (IDER), a downstream-aware task for histology-based spatial gene expression prediction. At evaluation time, for a specified biological or morphology-defined contrast, we compute gene-level differential-expression statistics from both measured and predicted expression profiles and evaluate whether the resulting contrast-specific ranked gene lists agree, as illustrated in Figure~\ref{fig:intro} (b). This criterion directly assesses whether predicted profiles preserve the gene prioritization used in DEG-based discovery, rather than only evaluating each gene's spatial pattern in isolation. Importantly, genes are not compared by raw expression magnitudes across genes; they are compared through their differential statistics, as in standard differential expression analysis.

Beyond evaluation, we introduce a training objective
specifically designed for IDER. During training, the objective uses morphology-derived proxy contrasts obtained by clustering histology features, without requiring predefined biological group labels such as tissue-region or phenotype annotations. For each proxy contrast, we compute differentiable differential-expression statistics from predicted profiles and align them with the corresponding statistics computed from measured profiles. This encourages the model to capture genes whose expression differences are associated with morphology-defined image variation, rather than only reconstructing individual spatial expression patterns.

We evaluate the proposed framework on public ST datasets under both morphology-derived contrasts and pathologist-annotated tissue-region contrasts. Experiments show that the proposed objective improves DEG-ranking agreement and pathway-enrichment overlap compared with conventional reconstruction objectives. These results suggest that image-based differential expression ranking provides a useful downstream-aware formulation for connecting histology-based ST prediction with morphology-associated gene discovery.

Our contributions are summarized as follows:
\begin{itemize}
    \item We identify a mismatch between conventional objectives and evaluation protocols for histology-based ST prediction and downstream gene discovery from ST profiles. Conventional criteria primarily measure spatial-profile agreement within each gene, whereas differential expression analysis depends on contrast-specific ranked gene lists that prioritize genes by between-group expression differences.

    \item We formulate image-based differential expression ranking (IDER), a downstream-aware task that evaluates whether predicted expression profiles preserve the contrast-specific ranked gene lists induced by measured expression profiles. For a specified biological or morphology-defined contrast, IDER compares gene rankings derived from differential-expression statistics computed from predicted and measured profiles.

    \item We introduce a differentiable IDER objective that aligns differential-expression statistics across genes using morphology-derived proxy contrasts from clustered histology features, without requiring predefined biological group labels during training. Experiments on public ST datasets show improved DEG-ranking agreement and pathway-enrichment overlap over conventional reconstruction objectives, including evaluations with morphology-derived groups and pathologist-annotated tissue regions.
\end{itemize}

\section{Related Work}
\noindent
{\bf Spatial Gene Expression Estimation from Histopathological Images.}
Predicting gene expression from histology images has emerged as a promising direction for bridging morphological and molecular information. A pioneering work in this area is ST-Net~\cite{he2020integrating}, which estimates gene expression from image patches using convolutional neural networks. Subsequent studies have explored architectures that capture both global and local context. In particular, Transformer-based models~\cite{pang2021leveraging} and graph convolutional networks (GCNs)~\cite{ganguly2025merge,jia2024thitogene,zeng2022spatial} have been introduced to model long-range spatial dependencies. Multi-scale modeling strategies have also been proposed to effectively combine local and global contexts~\cite{Wang2024M2ORTMR,chung2024accurate}.
Retrieval-based approaches~\cite{xie2023spatially,yang2023exemplar,dang2025hage} estimate gene expression by leveraging similarities between images and transcriptomic profiles in a joint embedding space, rather than directly regressing expression values. Methods based on diffusion models~\cite{zhu2025diffusion} and flow matching~\cite{huang2025scalable} have also been proposed, formulating the task as generating image-conditioned gene expression distributions.

\noindent
{\bf Ranking-based Objectives for Gene Expression Estimation.}
Recent studies have introduced ranking-based objectives for gene expression prediction. RankByGene~\cite{huang2024rankbygene} uses spatial ranking-based alignment to improve image and gene expression correspondence, while STRank~\cite{nishimura2025learning} models relative expression ordering across spatial spots. 
Both methods mainly focus on spatial-level relativity, i.e., expression rankings across spots.

While prior studies have advanced histology-based gene expression estimation, they mainly aim to reconstruct spatial trends across target genes and do not explicitly model the relative biological importance of genes. In contrast, we focus on gene-wise rankings derived from differential expression analysis, which are useful for discovering genes associated with image-derived features.

\noindent
{\bf Utilizing Biological Signals in Image and Gene Expression Data.}
Recent studies have increasingly emphasized higher-level biological signals, such as pathways, rather than relying solely on individual gene expression profiles. PEaRL~\cite{majumder2026pearl} incorporates pathway priors into histology-to-transcriptomics learning, whereas SurvPath~\cite{jaume2024modeling} and PathCLAST~\cite{noh2026pathclast} integrate pathway graphs to encode known biological relationships during model estimation. HistoPrism~\cite{Hu2026HistoPrism} further advocates pathway-level benchmarking to evaluate whether predicted transcriptomic profiles preserve functionally coherent biological programs. 

Motivated by recent studies, we extend the ST estimation task toward downstream biological analysis to better capture biologically meaningful signals.

\section{Preliminary: Spatial Gene Expression Prediction from Histology Images}

The objective of conventional gene expression estimation is to reconstruct spot-wise gene expression values for all target genes, while preserving spatial expression patterns across spots~\cite{he2020integrating, pang2021leveraging,chung2024accurate, shiku2026auxiliary}.

Formally, given a set of spot images $\mathcal{X} = \{\mathbf{x}_i\}_{i=1}^{N}$ and corresponding gene expression vectors $\mathcal{Y} = \{\mathbf{y}_i\}_{i=1}^{N}$, where $\mathbf{y}_i \in \mathbb{R}^{G}$, $N$ denotes the number of spots and $G$ the number of genes, existing methods learn a mapping
\[
f: \mathbf{x}_i \mapsto \hat{\mathbf{y}}_i,
\]
by optimizing reconstruction-based objectives such as Mean Squared Error (MSE)~\cite{he2020integrating, pang2021leveraging,chung2024accurate} or correlation-based metrics such as the Pearson Correlation Coefficient (PCC)~\cite{shiku2026auxiliary, yang2023exemplar, yang2024spatial}. PCC-based objectives measure, for each gene separately, the correlation between predicted and measured expression profiles across spots.

These objectives are natural for reconstructing spatial expression profiles for individual genes. However, they do not explicitly assess whether the predicted profiles preserve the contrast-specific ranked gene lists used in DEG-based gene discovery. This motivates the downstream-aware formulation introduced in the following section.

\section{Task Definition: Image-based Differential Expression Ranking}

We define image-based differential expression ranking (IDER) as the task of making contrast-specific ranked gene lists induced by predicted expression profiles agree with those induced by measured expression profiles. Given group labels for spots, differential expression analysis assigns each gene a contrast statistic and ranks genes by differential-expression evidence. The output of interest is therefore not the statistic for each gene in isolation, but the ordering of genes induced by these statistics.

Formally, let $\mathcal{X} = \{\mathbf{x}_i\}_{i=1}^{N}$ denote pathology images, $\mathcal{Y} = \{\mathbf{y}_i\}_{i=1}^{N}$ the corresponding measured expression profiles, and $\mathcal{L} = \{l_i\}_{i=1}^{N}$ the group labels, where $\mathbf{y}_i \in \mathbb{R}^{G}$ and $l_i \in \{1,\dots,K\}$. For each group $c$, we consider a one-versus-rest contrast:
\[
A_c = \{ i \mid l_i = c \}, \quad
B_c = \{ j \mid l_j \neq c \}.
\]

For each contrast $c$ and gene $g$, we use the one-sided Mann--Whitney U statistic~\cite{mann1947test,stuart2019comprehensive} as a rank-based differential-expression score. Intuitively, $U_{c,g}$ measures how often a spot in the target group $A_c$ has higher expression of gene $g$ than a spot in the reference group $B_c$; larger values indicate stronger evidence that gene $g$ is more highly expressed in $A_c$ relative to $B_c$. We compute this score as:
\[
U_{c, g} = \sum_{i \in A_c} \sum_{j \in B_c}
\left[
\mathbf{1}[y_{i,g} > y_{j,g}]
+ \frac{1}{2}\mathbf{1}[y_{i,g} = y_{j,g}]
\right],
\]
where $\mathbf{1}[\cdot]$ denotes the indicator function, and ties are assigned half weight.

The reference DEG output for contrast $c$ is the ranked gene list induced by these scores:
\[
\pi_c = \operatorname{argsort}^{\downarrow}_{g}\left(U_{c,g}\right),
\]
where $\operatorname{argsort}^{\downarrow}$ returns gene indices in descending order of $U_{c,g}$. Thus, $\pi_c$ prioritizes genes with stronger one-sided differential-expression evidence in group $A_c$ relative to group $B_c$. Given predicted expression profiles $\hat{\mathcal{Y}} = \{\hat{\mathbf{y}}_i\}_{i=1}^{N}$, we analogously compute $\hat{U}_{c,g}$ by replacing $y_{i,g}$ with $\hat{y}_{i,g}$ and obtain $\hat{\pi}_c = \operatorname{argsort}^{\downarrow}_{g}(\hat{U}_{c,g})$. The goal of IDER is to make the predicted ranked gene list $\hat{\pi}_c$ agree with the reference ranked gene list $\pi_c$ for each contrast.

\section{Method: Differentiable Optimization for IDER}

Based on the IDER task, our goal is to learn an image-to-expression predictor whose outputs preserve contrast-specific DEG rankings. Since the ranked gene list $\hat{\pi}_c$ is induced by pairwise comparisons and sorting, directly optimizing this ranking is not differentiable. We therefore optimize a differentiable surrogate that aligns gene-level differential-expression statistics computed from predicted expression profiles with those computed from measured expression profiles.

\noindent
{\bf Biological group-label-free training with morphology-derived proxy contrasts.}
During training, our objective does not require predefined biological group labels such as tissue-region or phenotype annotations. Instead, we obtain morphology-derived proxy groups by clustering pathology foundation-model~\cite{lu2024visual} embeddings with K-means, and construct one-versus-rest contrasts by varying the target group during optimization. This exposes the model to diverse morphology-associated expression differences and encourages it to preserve cross-gene differential-expression ordering across contrasts.

\noindent
{\bf Differentiable approximation of the U statistic.}
For each training contrast $c$, with groups $A_c$ and $B_c$ defined as in the task formulation, we approximate the one-sided Mann--Whitney comparison by replacing the hard indicator $\mathbf{1}[\hat{y}_{i,g} > \hat{y}_{j,g}]$ with a temperature-scaled sigmoid function. The differentiable statistic for gene $g$ is defined as:
\begin{equation}
\hat{U}_{c,g}
=
\sum_{i \in A_c}
\sum_{j \in B_c}
\sigma\!\left( \frac{\hat{y}_{i,g} - \hat{y}_{j,g}}{\tau} \right),
\end{equation}
where $\sigma(\cdot)$ denotes the sigmoid function and $\tau>0$ is a temperature parameter. This statistic smoothly relaxes the one-sided U statistic: as $\tau \to 0$, it approaches hard pairwise comparison, while larger $\tau$ yields smoother gradients for stable optimization. Because network outputs are continuous, exact ties are rare, so we omit the tie term used in the discrete statistic.

\noindent
{\bf DEG-ranking surrogate objective.}
For each contrast $c$, let $\mathbf{U}_c=(U_{c,1},\ldots,U_{c,G})$ be the reference U-statistic vector and $\hat{\mathbf{U}}_c=(\hat{U}_{c,1},\ldots,\hat{U}_{c,G})$ be its differentiable counterpart computed from predicted profiles. The DEG ranking is determined by the relative magnitudes of these entries across genes. We therefore optimize a correlation-based surrogate over the gene axis:
\begin{equation}
\mathcal{L}_{\mathrm{DEG}}
=
1 -
\frac{1}{|\mathcal{C}|}
\sum_{c \in \mathcal{C}}
\mathrm{Corr}_{g}\left(\hat{\mathbf{U}}_c, \mathbf{U}_c\right),
\end{equation}
where $\mathrm{Corr}_{g}$ denotes the Pearson correlation computed across genes, and $\mathcal{C}$ denotes the set of group labels. 
This differs from conventional PCC losses, which compute correlation across spots for each fixed gene. 
Here, each gene is summarized by a contrast-specific differential-expression score, and the loss encourages genes with high reference U statistics to also receive high predicted U statistics, serving as a differentiable surrogate for preserving the contrast-specific DEG ranking.

\noindent
{\bf Auxiliary expression-consistency regularization.}
The DEG-ranking objective aligns differential-expression evidence across genes, but it does not explicitly constrain the spatial expression profile of each individual gene. We therefore add a lightweight expression-consistency regularizer $\mathcal{L}_{\mathrm{const}}$, defined as a conventional PCC-based loss that correlates predicted and measured expression across spots for each gene and averages the result over genes. This term is used to stabilize training and preserve standard spatial-profile agreement, while the main objective remains DEG-ranking reproducibility. The overall loss is:
\begin{equation}
\mathcal{L} = \mathcal{L}_{\mathrm{DEG}} + \lambda \mathcal{L}_{\mathrm{const}},
\end{equation}
where $\lambda$ controls the strength of the auxiliary regularization.

\section{Evaluation Metrics for IDER}

We evaluate whether predicted profiles preserve the contrast-specific DEG output using two complementary ranking metrics. For each contrast $c$, DEG-SCC measures global agreement of differential-expression evidence by computing Spearman's rank correlation across genes $\mathrm{Spearman}_g(\cdot)$ between the reference and predicted U-statistic vectors:
\begin{equation}
\mathrm{SCC}^{\mathrm{DEG}}_c
=
\mathrm{Spearman}_g
\left(
\mathbf{U}_c,
\hat{\mathbf{U}}_{c}
\right).
\end{equation}

To emphasize recovery of top-ranked DEG candidates, we additionally use binary DEG-nDCG@$k$~\cite{jarvelin2002cumulated}. Let $\pi_c(r)$ and $\hat{\pi}_c(r)$ denote the genes ranked at position $r$ in the reference and predicted DEG lists, respectively. We define the binary relevance of gene $g$, indicating whether the gene is included in the top-$k$ genes of the reference DEG list, as
$\mathrm{rel}_c(g)
=
\mathbf{1}\left[g \in \{\pi_c(1),\ldots,\pi_c(k)\}\right],
$
where $\mathbf{1}[\cdot]$ is the indicator function. Then, binary nDCG@$k$ is defined as
\begin{equation}
\mathrm{nDCG}^{\mathrm{DEG}}_c@k
=
\frac{
\sum_{r=1}^{k}
\frac{\mathrm{rel}_c(\hat{\pi}_c(r))}{\log_2(r+1)}
}{
\sum_{r=1}^{k}
\frac{1}{\log_2(r+1)}
}.
\end{equation}

$\mathrm{SCC}^{\mathrm{DEG}}_c$ evaluates agreement across all genes, whereas $\mathrm{nDCG}^{\mathrm{DEG}}_c$@$k$ focuses on whether genes highly ranked in the reference DEG list also appear near the top of the predicted DEG list. We report both metrics averaged across the considered contrasts.

\section{Experiments}
We evaluate the proposed method on spatial transcriptomics datasets to assess its ability to estimate DEG rankings from inter-group comparisons.  
We consider two settings: (i) a \textbf{cluster-based DEG setting} (Sec. \ref{sec:cluster}), where groups are defined via clustering of pathology foundation model features, and (ii) a \textbf{real annotation-based DEG setting} (Sec. \ref{sec:annot}), where groups correspond to pathologist-provided tissue regions.  
Across both settings, we further analyze downstream pathway recovery, compare against conventional objectives, and conduct ablation and plug-in studies (Sec. \ref{sec:analy}).

\noindent
{\bf Implementation Details.}
All experiments are implemented in PyTorch~\cite{Paszke2019PyTorchAI} and optimized using Adam~\cite{adam} with a learning rate of $3 \times 10^{-4}$.  
We use CONCH~\cite{lu2024visual} as the feature extractor for the gene expression prediction network $f$, with its parameters frozen during training. Only a three-layer MLP following the feature extractor is trained.
Models are trained for 1,000 epochs, and we report the best validation checkpoint.
The auxiliary regularization weight and temperature parameter are set to $\lambda = 1.0$ and $\tau = 0.01$, respectively.

During training, groups are defined as regions sharing similar cellular morphology, obtained via K-means clustering on image features with the number of clusters set to 10.  
We adopt a one-vs-rest scheme, where a target group is treated as positive and the others as negative. The positive group is randomly sampled at each iteration, encouraging the model to learn DEG rankings across all groups.

\noindent
{\bf Evaluation Metrics.}
We use evaluation metrics based on group-wise comparison that assess the ranking of DEGs obtained by the proposed framework.
Evaluation is conducted using SCC$^{\scriptsize \mathrm{DEG}}$, which reflects the global agreement of DEG rankings, and $\mathrm{nDCG}^{\mathrm{DEG}}$@$k$, which evaluates the recovery of highly ranked genes. We report $\mathrm{nDCG}^{\mathrm{DEG}}$@50, @100, and @200.
Similar to training, we construct a one-vs-rest setting in which one group is treated as the positive class and the remaining groups as negative classes, and compute the ranking of DEGs for evaluation.
This procedure is performed for all groups, and the final performance is obtained by averaging the results across groups.

\noindent
{\bf Baseline Objective Functions.}
To demonstrate the effectiveness of the proposed DEG-based ranking objective, we compare our method with six conventional baselines that optimize only spatial expression independently for each gene.
1) to 3) ``MSE''~\cite{he2020integrating, pang2021leveraging, chung2024accurate}, ``PCC''~\cite{shiku2026auxiliary}, and ``MSE $\&$ PCC''~\cite{yang2023exemplar, yang2024spatial} are the most commonly used objective functions in state-of-the-art ST estimation methods, where the model is trained using losses computed independently for each gene, such as MSE loss, PCC loss, or their combination.
4) ``Poisson'' and 5) ``NB''~\cite{nishimura2026cell} are likelihood-based objectives for modeling gene expression counts at the individual sample level, where Poisson assumes a Poisson distribution and NB additionally models overdispersion using a dispersion parameter.
6) ``STRank''~\cite{nishimura2025learning} is a state-of-the-art objective function that learns relative spatial expression patterns through a ranking-based loss robust to stochastic noise in ST measurements.

\subsection{Cluster-based DEG Experiments} \label{sec:cluster}
\noindent
{\bf Datasets.}
We conducted experiments on the Bowel, Ovary, and Lymph Node samples from the public HEST-1K dataset~\cite{jaume2024hest}. 
For each organ, we used a single slide to enable a controlled comparison of different training objectives without the influence of cross-patient batch effects.
Each dataset consists of paired patch images and gene expression profiles, with 4,096, 4,617, and 4,840 samples for Bowel, Ovary, and Lymph Node, respectively. All patch images are of size $224 \times 224$.
ST data were obtained using Visium technology~\cite{williams2022introduction}. 
After preprocessing, as detailed in the Supplementary Material (Sec.~A.7),
the numbers of genes were 18,066, 18,054, and 17,564
for Bowel, Ovary, and Lymph Node, respectively.
While conventional studies typically select only a small subset of highly variable genes (HVGs) for training and evaluation~\cite{jaume2024hest, nishimura2025learning}, our goal is to evaluate DEG rankings across all genes; therefore, we use all genes in our experiments.
Following~\cite{shiku2026auxiliary}, we split each dataset into training, validation, and test sets with a ratio of 3:1:1.

{
\renewcommand{\arraystretch}{1.2}
\begin{table}[t]
    \centering
    \caption{\textbf{Comparison of DEG ranking performance in the cluster-based DEG setting} against methods optimized with conventional objective functions.}
    \label{tab:comparison_cluster}
    \scalebox{0.71}{
    \begin{tabular}{c|cccc|cccc|cccc}
        \toprule
        \multirow{3}{*}{Objective}
        & \multicolumn{4}{c|}{\textbf{Ovary}}
        & \multicolumn{4}{c|}{\textbf{Lymph Node}} 
        & \multicolumn{4}{c}{\textbf{Bowel}}\\
        
        \cmidrule(lr){2-5} \cmidrule(lr){6-9} \cmidrule(lr){10-13}
        
        & \multirow{2}{*}{SCC$^{\scriptsize \mathrm{DEG}}$}
        & \multicolumn{3}{c|}{$\mathrm{nDCG}^{\mathrm{DEG}}$}
        & \multirow{2}{*}{SCC$^{\scriptsize \mathrm{DEG}}$}
        & \multicolumn{3}{c|}
        {$\mathrm{nDCG}^{\mathrm{DEG}}$}
        & \multirow{2}{*}{SCC$^{\scriptsize \mathrm{DEG}}$}
        & \multicolumn{3}{c}{$\mathrm{nDCG}^{\mathrm{DEG}}$} \\
        
        \cmidrule(lr){3-5}
        \cmidrule(lr){7-9}
        \cmidrule(lr){11-13}
        
        &  & @50 & @100 & @200
        &  & @50 & @100 & @200
        &  & @50 & @100 & @200 \\
        
        \midrule
        MSE~\cite{chung2024accurate}          & 0.587 & 0.415 & 0.446 & 0.434
                     & 0.624 & 0.407 & 0.434 & 0.477
                     & 0.709 & 0.364 & 0.432 & 0.468\\
        
        PCC~\cite{shiku2026auxiliary}           & 0.584 & 0.425 & 0.451 & 0.451
                     & 0.629 & 0.439 & 0.478 & 0.521
                     & 0.710 & 0.384 & 0.461 & 0.494\\
        
        MSE $\&$ PCC~\cite{yang2023exemplar}   & 0.609 & 0.437 & 0.465 & 0.453
                     & 0.650 & 0.477 & 0.511 & 0.548
                     & 0.714 & 0.383 & 0.459 & 0.496
                     \\
        
        Poisson      & 0.460 & 0.363 & 0.401 & 0.422
                     &0.234& 0.448 & 0.493 & 0.501
                     & 0.694 & 0.440 & 0.476 & 0.492
                     \\
        
        NB~\cite{nishimura2026cell}           & 0.245 & 0.404 & 0.415 & 0.417
                     & 0.090 & 0.541 & 0.548 & 0.534
                     & 0.601 & \textbf{0.494} & 0.533 & 0.546
                     \\
        
        STRank~\cite{nishimura2025learning}      & 0.457 & 0.359 & 0.399 & 0.416
                     & 0.300 & 0.375 & 0.425 & 0.446
                     & 0.687 & 0.435 & 0.470 & 0.498\\
        
        \rowcolor{gray!15}
        \textbf{Ours}
                     & \textbf{0.686} & \textbf{0.486} & \textbf{0.515} & \textbf{0.531}
                     & \textbf{0.676} & \textbf{0.601} & \textbf{0.641} & \textbf{0.643}
                     & \textbf{0.731} & 0.481 & \textbf{0.565} & \textbf{0.595}
                     \\
        
        \bottomrule
    \end{tabular}
    }
    \vspace{-0.5mm}
\end{table}

These datasets do not include annotations for tissue regions such as diseased or normal areas.
Therefore, we construct groups of tissues with similar morphological characteristics by clustering features extracted using the pretrained pathology foundation model CONCH~\cite{lu2024visual}.
\begin{wrapfigure}{r}{0.5\linewidth}
  \centering
  \vspace{-1pt}
  \includegraphics[width=0.92\linewidth]{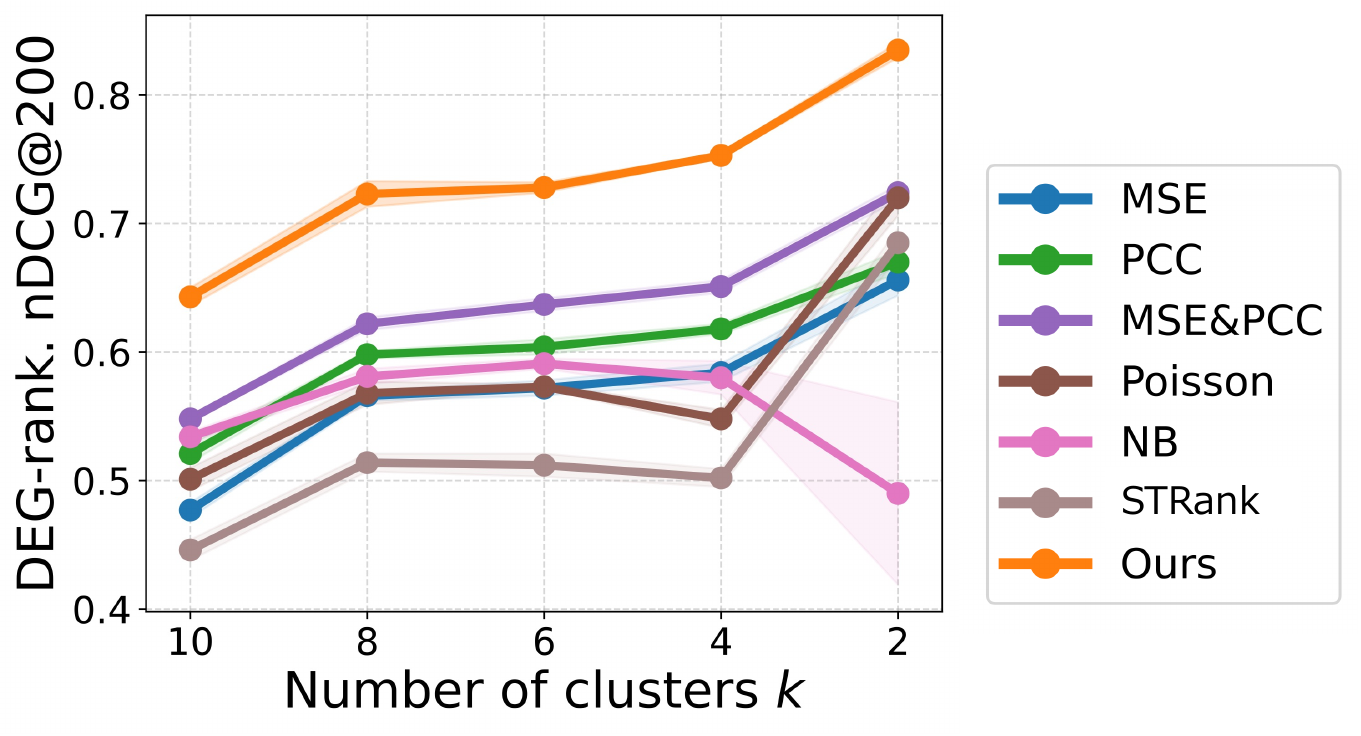}
  \caption{\textbf{Generalization Performance Across Various Grouping Granularities.}}
    \label{fig:robustness_group_size}
  \vspace{-10pt}
\end{wrapfigure}
The clustering procedure is performed completely independently on the training data to ensure that no data leakage occurs.
This approach allows us to generate groups at arbitrary granularity simply by varying the number of clusters, without requiring costly annotation procedures.
Thus, this setting is not only useful for evaluating machine learning methods, but also has the potential to facilitate efficient differential expression analysis in real-world gene expression studies by enabling annotation-free grouping.

\noindent
{\bf Comparison of DEG Ranking Performance.}
Table~\ref{tab:comparison_cluster} reports DEG ranking results under the cluster-based DEG setting.
The proposed method achieves the best performance on nearly all metrics, demonstrating substantial improvements over conventional objectives.
These results suggest that conventional methods, which optimize the reproducibility of gene-wise spatial expression patterns, do not necessarily improve DEG ranking estimation between morphologically distinct tissue groups.
The strong performance of our method further highlights the importance of explicitly optimizing DEG rankings derived from morphology-based proxy groups.

{\bf Evaluation of Generalization Performance Across Various Grouping Granularities.}
In practical applications, the granularity of group partitioning varies depending on the objective of differential expression analysis, making it important to generalize DEG estimation across different grouping granularities. Therefore, in this experiment, we evaluate the generalization capability of the proposed method under varying grouping granularities using the Lymph Node dataset.

Figure~\ref{fig:robustness_group_size} shows the DEG ranking performance of each method, evaluated by $\mathrm{nDCG}^{\mathrm{DEG}}$@200, under the cluster-based DEG setting with varying numbers of clusters $k=10, 8, 6, 4, 2$.
The proposed method consistently achieves the best performance across different grouping granularities. These results indicate that, although trained with a specific grouping granularity, the proposed method generalizes well to DEG ranking estimation under different grouping granularities, demonstrating strong potential for practical differential expression analyses at various levels of granularity.

\subsection{Real Annotation-based DEG Experiments}
\label{sec:annot}
\noindent
{\bf Datasets.} 
We conduct experiments using the publicly available breast cancer ST dataset Her2st~\cite{andersson2020spatial}, which includes expert-provided region annotations.
Her2st contains histopathology images and ST data acquired from 8 patients using Spatial Transcriptomics technology. For each patient, 3 to 6 histopathology images are available, resulting in a total of 36 images.
After applying the preprocessing procedure detailed in the
Supplementary Material (Sec.~A.7), we obtain 13,620 pairs of
$224 \times 224$ image patches and gene expression profiles
with 11,871 genes.
As in the previous setting, all genes are used for both training and evaluation.
We conduct patient-level cross-validation following~\cite{chung2024accurate} by splitting the dataset at the patient level with a ratio of 2:1:1.

In this dataset, one WSI per patient is annotated with tissue-type region labels among the multiple WSIs available for each patient.
The annotations include adipose tissue, cancer in situ, connective tissue, immune infiltrate, invasive cancer, breast glands, and undetermined.
For evaluation, we use annotated WSIs from the test patients and treat each tissue type as a distinct group. 
We exclude the undetermined category, as well as cancer in situ and breast glands, which are present in only a small subset of patients. In addition, classes with fewer than 10 spots per patient are excluded from evaluation, as reliable statistical testing is not feasible.
Importantly, we use the annotations only for evaluation and not for training at all. Instead, training is performed using groups obtained through the automatic group construction method proposed in our framework.

\noindent
{\bf Comparison of DEG Ranking Performance.} Table~\ref{tab:annot_comparison} presents the DEG ranking performance for all subtypes as well as for each individual subtype under the real annotation-based DEG setting.
Due to space limitations, for $\mathrm{nDCG}^{\mathrm{DEG}}$@$k$, we report only the results for @200.
The proposed method consistently outperformed baseline methods across all tissue types and evaluation metrics.
Similar to the cluster-based setting, it improved both the global agreement of DEG rankings measured by SCC$^{\scriptsize \mathrm{DEG}}$ and the identification of top-ranked genes measured by $\mathrm{nDCG}^{\mathrm{DEG}}$@200.
Notably, although the model was trained without tissue-type annotations and relied only on clusters derived from image features, it generalized well to real pathological annotations at test time.
These results suggest that the proposed objective encourages the model to capture gene signals more strongly associated with histopathological tissue patterns.
Therefore, compared with existing approaches, the proposed method may provide a more effective framework for discovering genes associated with image-derived pathological findings.

\subsection{Further Analysis and Validation}
\label{sec:analy}

\noindent
\begin{wraptable}{r}{0.52\linewidth}
    \centering
    \vspace{-4.5mm}
    \caption{\textbf{Analysis of pathway enrichment overlap.}}
    \label{tab:pathway}
    \scalebox{0.78}{
    \renewcommand{\arraystretch}{1.2}
    \begin{tabular}{c|cccc}
        \toprule
        Objective  & Ovary & Lymph Node & Bowel & Breast \\
        \midrule
        MSE~\cite{chung2024accurate}         & 0.326 & 0.382 & 0.336 & 0.144 \\
        PCC~\cite{shiku2026auxiliary}         & 0.341 & 0.414 & 0.345 & 0.127 \\
        MSE $\&$ PCC~\cite{yang2023exemplar} & 0.332 & 0.433 & 0.314 & 0.125 \\
        Poisson     & 0.307 & 0.413 & 0.362 & 0.136 \\
        NB~\cite{nishimura2026cell}          & 0.289 & 0.443 & 0.385 & 0.143 \\
        STRank~\cite{nishimura2025learning}      & 0.302 & 0.383 & 0.360 & 0.147 \\
        \rowcolor{gray!15}
        \textbf{Ours}
                    & \textbf{0.381} & \textbf{0.571} & \textbf{0.405} & \textbf{0.235} \\
        \bottomrule
    \end{tabular}
    \vspace{-2mm}
    }
\end{wraptable}
{\bf Pathway Enrichment Overlap Analysis for Estimated ST.}
To assess whether predicted profiles preserve downstream functional interpretation, we perform pathway enrichment analysis using the top-ranked genes from the predicted and measured DEG lists. 
For each contrast, we first select the top 200 genes from the predicted and measured DEG-ranked lists, respectively. We then conduct enrichment analysis for each gene set using the Reactome\_2022 and KEGG\_2021\_Human pathway panels~\cite{gillespie2022reactome,kanehisa2021kegg}, and retain the top 200 enriched pathways according to the enrichment scores. 
Subsequently, following previous studies~\cite{Maleki2019MethodCI,kerseviciute2023apear}, we compute the Jaccard index between the predicted and measured pathway sets to quantify their overlap.
This analysis evaluates whether the proposed method preserves pathway-level interpretations derived from contrast-specific gene prioritization, rather than only matching individual gene expression profiles. We conduct this analysis on all datasets under both morphology-derived group and pathologist-annotated tissue-region settings.

Table~\ref{tab:pathway} shows that the proposed method achieved the highest Jaccard scores across all datasets, indicating that it better recovered biologically relevant pathways compared with baseline methods.
These results suggest that the proposed method better preserves biological functional structures in ST and more accurately estimates biologically meaningful signals from pathology images.

\begin{table}[t]
    \centering
    \caption{\textbf{Comparison of DEG ranking performance in the real annotation-based DEG setting} against methods optimized with conventional objective functions, in terms of SCC$^{\scriptsize \mathrm{DEG}}$ and $\mathrm{nDCG}^{\mathrm{DEG}}$@200. Due to space limitations, $\mathrm{nDCG}^{\mathrm{DEG}}$@200 is abbreviated as $\mathrm{nDCG}^{\mathrm{DEG}}$.}
    \scalebox{0.635}{
    \begin{tabular}{c|cc|cc|cc|cc|cc}
        \toprule
        \multirow{2}{*}{Objective} 
        & \multicolumn{2}{c|}{\textbf{Average}} 
        & \multicolumn{2}{c|}{\textbf{adipose tissue}}  
        & \multicolumn{2}{c|}{\textbf{connective tissue}} 
        & \multicolumn{2}{c|}{\textbf{immune infiltrate}} 
        & \multicolumn{2}{c}{\textbf{invasive cancer}} \\  
        \cmidrule(lr){2-3} \cmidrule(lr){4-5} \cmidrule(lr){6-7} \cmidrule(lr){8-9} \cmidrule(lr){10-11}
        & SCC$^{\scriptsize \mathrm{DEG}}$ & $\mathrm{nDCG}^{\mathrm{DEG}}$
        & SCC$^{\scriptsize \mathrm{DEG}}$ & $\mathrm{nDCG}^{\mathrm{DEG}}$
        & SCC$^{\scriptsize \mathrm{DEG}}$ & $\mathrm{nDCG}^{\mathrm{DEG}}$
        & SCC$^{\scriptsize \mathrm{DEG}}$ & $\mathrm{nDCG}^{\mathrm{DEG}}$
        & SCC$^{\scriptsize \mathrm{DEG}}$ & $\mathrm{nDCG}^{\mathrm{DEG}}$ \\
        \midrule
        MSE~\cite{chung2024accurate}          & 0.239 & 0.084 & 0.393 & 0.069 & 0.206 & 0.080 & 0.197 & 0.176 & 0.272 & 0.082 \\
        PCC~\cite{shiku2026auxiliary}          & 0.181 & 0.073 & 0.297 & 0.060 & 0.143 & 0.066 & 0.132 & 0.199 & 0.182 & 0.035 \\
        MSE $\&$ PCC~\cite{yang2023exemplar} & 0.195 & 0.073 & 0.317 & 0.076 & 0.208 & 0.048 & 0.134 & 0.239 & 0.216 & 0.043 \\
        Poisson      & 0.116 & 0.101 & 0.123 & 0.057 & 0.105 & 0.109 & 0.191 & 0.228 & 0.114 & 0.081 \\
        NB~\cite{nishimura2026cell}           & 0.137 & 0.098 & 0.145 & 0.038 & 0.127 & 0.097 & 0.190 & 0.269 & 0.120 & 0.073 \\
        STRank~\cite{nishimura2025learning}      & 0.101 & 0.109 & 0.053 & 0.083 & 0.114 & 0.121 & 0.189 & 0.227 & 0.096 & 0.077 \\
        \rowcolor{gray!15}
        \textbf{Ours}
                     & \textbf{0.353} & \textbf{0.247}
                     & \textbf{0.475} & \textbf{0.100}
                     & \textbf{0.313} & \textbf{0.164}
                     & \textbf{0.263} & \textbf{0.446}
                     & \textbf{0.357} & \textbf{0.301} \\
        \bottomrule
    \end{tabular}
    }
    \label{tab:annot_comparison}
\end{table}

\noindent
{\bf Ablation Experiments.}
We conduct ablation experiments to analyze the contribution of each component in the proposed method.
We compare the following two ablation methods with our proposed method: (i) ``Baseline,'' a baseline model trained with a PCC loss that learns spatial expression patterns for each gene independently; and (ii) ``Ours w/o DEG ranking,'' which additionally introduces a gene-independent MSE loss on the DEG-based statistic $U$ without considering ranking relationships between DEGs.
These experiments were conducted on all datasets, and performance was evaluated using SCC$^{\scriptsize \mathrm{DEG}}$ and $\mathrm{nDCG}^{\mathrm{DEG}}$@200.

Table~\ref{tab:comparison_ablation} shows the ablation results, where $\mathrm{nDCG}^{\mathrm{DEG}}$ denotes $\mathrm{nDCG}^{\mathrm{DEG}}$@200 due to space limitations.
Baseline achieved reasonable SCC\(^{\mathrm{DEG}}\) performance but showed limited ability to identify top-ranked genes, as reflected by the lower $\mathrm{nDCG}^{\mathrm{DEG}}$@200 scores.
Introducing MSE-based supervision on \(U\) in ``Ours w/o DEG ranking'' partially improved $\mathrm{nDCG}^{\mathrm{DEG}}$@200, suggesting the effectiveness of incorporating DEG signals.
However, the improvements were inconsistent, indicating that gene-independent supervision on U alone is insufficient to capture inter-gene ranking relationships.
In contrast, the proposed method, which optimizes inter-gene ranking relationships on U, achieved substantial improvements across all metrics.
These results highlight the importance of optimizing inter-gene ranking relationships for DEG ranking estimation.   

\begin{table}[t]
    \centering
    \caption{\textbf{Ablation experiments of the proposed method}.}
    \scalebox{0.7}{
    \setlength{\tabcolsep}{3pt}
    \begin{tabular}{c|cc|cc|cc|cc|cc}
        \toprule
        \multirow{3}{*}{Objective} &&
        & \multicolumn{2}{c|}{\textbf{Ovary}}
        & \multicolumn{2}{c|}{\textbf{Lymph Node}} 
        & \multicolumn{2}{c|}{\textbf{Bowel}}
        & \multicolumn{2}{c}{\textbf{Her2st}}\\
        
        \cmidrule(lr){4-5}
        \cmidrule(lr){6-7}
        \cmidrule(lr){8-9}
        \cmidrule(lr){10-11}

        &U stat.& DEG rank.&
        SCC$^{\scriptsize \mathrm{DEG}}$ & $\mathrm{nDCG}^{\mathrm{DEG}}$
        & SCC$^{\scriptsize \mathrm{DEG}}$ & $\mathrm{nDCG}^{\mathrm{DEG}}$
        & SCC$^{\scriptsize \mathrm{DEG}}$ & $\mathrm{nDCG}^{\mathrm{DEG}}$
        & SCC$^{\scriptsize \mathrm{DEG}}$ & $\mathrm{nDCG}^{\mathrm{DEG}}$\\
        
        \midrule        
        Baseline 
        &\ding{55} &\ding{55}
        & 0.584 & 0.451
        & 0.629 & 0.521
        & 0.710 & 0.494
        & 0.181 & 0.073\\
        
        Ours w/o DEG rank.
        &\ding{51} &\ding{55}
        & 0.587 & 0.454
        & 0.617 & 0.521
        & 0.710 & 0.491
        & 0.178 & 0.077\\
        
        \rowcolor{gray!15}
        \textbf{Ours}
        &\ding{51} &\ding{51}
        & \textbf{0.686} & \textbf{0.531}
        & \textbf{0.676} & \textbf{0.643}
        & \textbf{0.731} & \textbf{0.595}
        & \textbf{0.353} & \textbf{0.247}\\
        
        \bottomrule
    \end{tabular}
    }
    \label{tab:comparison_ablation}
    \vspace{-2mm}
\end{table}

\begin{figure}[t]
\centering

\begin{minipage}[t]{0.45\columnwidth}
\centering
\captionof{table}{\textbf{Evaluation on per-gene spatial-profile PCC.} L. N. indicates Lymph Node.}
\label{tab:comparison_conv_pearson}

{\small
\scalebox{0.8}{
\begin{tabular}{c|cccc}
\toprule
Objective  & Ovary & L. N. & Bowel & Breast \\
\midrule
MSE~\cite{chung2024accurate} & \textbf{0.235} & 0.171 & \textbf{0.345} & 0.113\\
PCC~\cite{shiku2026auxiliary} & 0.233 & \textbf{0.172} & \textbf{0.345} & \textbf{0.145}\\
MSE $\&$ PCC~\cite{yang2023exemplar} & \textbf{0.235} & \textbf{0.172} & 0.341 & 0.141\\
Poisson & 0.192 & 0.126 & 0.300 & 0.052\\
NB~\cite{nishimura2026cell} & 0.195 & 0.130 & 0.303 & 0.056\\
STRank~\cite{nishimura2025learning} & 0.190 & 0.130 & 0.300 & 0.065\\
\rowcolor{gray!15}\textbf{Ours}
& 0.216 & 0.148 & 0.322 & 0.097\\
\bottomrule
\end{tabular}
}}
\end{minipage}
\hfill
\begin{minipage}[t]{0.5\columnwidth}
\centering
\captionof{table}{\textbf{Plug-in experiments} for existing methods on the Her2st dataset.}
\label{tab:plugin}
    \vspace{2mm}

{\small
\setlength{\tabcolsep}{3pt}
\scalebox{0.83}{
\begin{tabular}{c|cccc}
\toprule
\multirow{2}{*}{Objective} &  & \multicolumn{3}{c}{$\mathrm{nDCG}^{\mathrm{DEG}}$} \\
\cmidrule(lr){3-5}
& SCC$^{\scriptsize \mathrm{DEG}}$ & @50 & @100 & @200 \\
\midrule
ST-Net~\cite{he2020integrating} & 0.181 & 0.060 & 0.062 & 0.073 \\
\rowcolor{gray!15}
ST-Net + \textbf{Ours} & \textbf{0.353} & \textbf{0.212} & \textbf{0.230} & \textbf{0.247} \\
\cmidrule(lr){1-5}
TRIPLEX~\cite{chung2024accurate} & 0.299 & 0.079 & 0.097 & 0.115\\
\rowcolor{gray!15}
TRIPLEX+ \textbf{Ours} & \textbf{0.457} & \textbf{0.190} & \textbf{0.212} & \textbf{0.242}\\
\bottomrule
\end{tabular}
}}
\end{minipage}
\vspace{-3mm}

\end{figure}

\noindent
{\bf Performance in Estimating Per-Gene Spatial Expression Patterns.}
We also evaluate all methods using conventional per-gene PCC, which measures spatial-profile agreement across spots. As shown in Table~\ref{tab:comparison_conv_pearson}, MSE- and PCC-based objectives generally achieve the highest scores, as expected because this metric matches their training targets. Although our method is not best under this reconstruction-oriented metric, it remains in an intermediate range, suggesting that the auxiliary expression-consistency term preserves reasonable spatial-profile agreement.

More importantly, together with the DEG-ranking and pathway-overlap results, this analysis highlights the gap between conventional reconstruction evaluation and downstream gene-discovery performance. Methods with higher per-gene spatial PCC do not necessarily achieve better DEG-ranking agreement or pathway-enrichment overlap, whereas our method improves these downstream metrics despite not being optimized primarily for conventional PCC. These results support the need to evaluate histology-based ST prediction not only by spatial-profile reconstruction, but also by whether predicted profiles preserve contrast-specific gene prioritization for downstream analysis.

\noindent
{\bf Plug-in to Existing Methods.}
The proposed method is model-architecture-agnostic and can therefore be integrated into existing ST prediction architectures.
Here, we applied the proposed method as a plug-in to two representative ST prediction methods, ST-Net~\cite{he2020integrating} and TRIPLEX~\cite{chung2024accurate}, and evaluated them on the Her2st dataset.
The results shown in Table~\ref{tab:plugin} demonstrate that DEG ranking performance was consistently and substantially improved for both methods.
This indicates that the proposed method can be easily applied in a plug-and-play manner and has strong generalizability.



\section{Conclusion}
In this paper, we highlighted the mismatch between conventional histology-based ST prediction based on per-gene spatial-profile reconstruction and differential expression analysis, an important downstream biological analysis that ranks genes by between-group expression differences. To address this issue, we proposed Image-based Differential Expression Ranking (IDER), a framework that evaluates whether predicted expression profiles preserve contrast-specific DEG rankings. We further introduced a differentiable objective that aligns differential-expression statistics across genes using morphology-derived proxy groups without predefined biological annotations. Experiments on public ST datasets demonstrated that the proposed method improved DEG-ranking agreement and pathway-enrichment overlap compared with conventional reconstruction objectives.

\begin{ack}
This work was supported by JST BOOST Grant Number JPMJBS2406,
SIP Grant Number JPJ012425,
JST ASPIRE Grant Number JPMJAP2403,
JSPS KAKENHI Grant Numbers JP26K21391 and JP24KJ2205,
and MEXT Supporting Pioneering Research through AI for 1,000
Discovery Challenges Program, SPReAD, Japan Grant Number JPMXP1726260906.
\end{ack}

\bibliographystyle{plain}  
\bibliography{refs}  

\begin{thebibliography}{10}

\bibitem{Paszke2019PyTorchAI}
Paszke Adam, Gross Sam, Massa Francisco, Lerer Adam, Bradbury James, Chanan Gregory, Killeen Trevor, Lin Zeming, Gimelshein Natalia, Antiga Luca, Desmaison Alban, K{\"o}pf Andreas, Yang Edward, DeVito Zach, Raison Martin, Tejani Alykhan, Chilamkurthy Sasank, Steiner Benoit, Fang Lu, Bai Junjie, and Chintala Soumith.
\newblock {PyTorch: An Imperative Style, High-Performance Deep Learning Library}.
\newblock In {\em Neural Information Processing Systems}, pages 8026 -- 8037, 2019.

\bibitem{andersson2020spatial}
Alma Andersson, Ludvig Larsson, Linnea Stenbeck, Fredrik Salm{\'e}n, Anna Ehinger, Sunny Wu, Ghamdan Al-Eryani, Daniel Roden, Alex Swarbrick, {\AA}ke Borg, et~al.
\newblock {Spatial Deconvolution of HER2-positive Breast Tumors Reveals Novel Intercellular Relationships}.
\newblock {\em BioRxiv}, pages 2020--07, 2020.

\bibitem{cable2022cell}
Dylan~M Cable, Evan Murray, Vignesh Shanmugam, Simon Zhang, Luli~S Zou, Michael Diao, Haiqi Chen, Evan~Z Macosko, Rafael~A Irizarry, and Fei Chen.
\newblock {Cell Type-Specific Inference of Differential Expression in Spatial Transcriptomics}.
\newblock {\em Nature Methods}, 19(9):1076--1087, 2022.

\bibitem{chen2024towards}
Richard~J Chen, Tong Ding, Ming~Y Lu, Drew~FK Williamson, Guillaume Jaume, Andrew~H Song, Bowen Chen, Andrew Zhang, Daniel Shao, Muhammad Shaban, et~al.
\newblock {Towards a General-purpose Foundation Model for Computational Pathology}.
\newblock {\em Nature Medicine}, 30(3):850--862, 2024.

\bibitem{chen2025visual}
Weiqing Chen, Pengzhi Zhang, Tu~N Tran, Yiwei Xiao, Shengyu Li, Vrutant~V Shah, Hao Cheng, Kristopher~W Brannan, Keith Youker, Li~Lai, et~al.
\newblock {A Visual--omics Foundation Model to Bridge Histopathology with Spatial Transcriptomics}.
\newblock {\em Nature Methods}, 22(7):1568--1582, 2025.

\bibitem{chung2024accurate}
Youngmin Chung, Ji~Hun Ha, Kyeong~Chan Im, and Joo~Sang Lee.
\newblock {Accurate Spatial Gene Expression Prediction by Integrating Multi-resolution Features}.
\newblock In {\em Computer Vision and Pattern Recognition}, pages 11591--11600, 2024.

\bibitem{dang2025hage}
Thao~M Dang, Haiqing Li, Yuzhi Guo, Hehuan Ma, Feng Jiang, Yuwei Miao, Qifeng Zhou, Jean Gao, and Junzhou Huang.
\newblock {HAGE: Hierarchical Alignment Gene-Enhanced Pathology Representation Learning with Spatial Transcriptomics}.
\newblock In {\em International Conference on Medical Image Computing and Computer-Assisted Intervention}, pages 228--238, 2025.

\bibitem{ganguly2025merge}
Aniruddha Ganguly, Debolina Chatterjee, Wentao Huang, Jie Zhang, Alisa Yurovsky, Travis~Steele Johnson, and Chao Chen.
\newblock {Merge: Multi-faceted Hierarchical Graph-based GNN for Gene Expression Prediction from Whole Slide Histopathology Images}.
\newblock In {\em Computer Vision and Pattern Recognition Conference}, pages 15611--15620, 2025.

\bibitem{gillespie2022reactome}
Marc Gillespie, Bijay Jassal, Ralf Stephan, Marija Milacic, Karen Rothfels, Alexandra Senff-Ribeiro, Johannes Griss, Cristoffer Sevilla, Lisa Matthews, Chuqiao Gong, et~al.
\newblock {The Reactome Pathway Knowledgebase 2022}.
\newblock {\em Nucleic Acids Research}, 50(D1):D687--D692, 2022.

\bibitem{he2020integrating}
Bryan He, Ludvig Bergenstr{\aa}hle, Linnea Stenbeck, Abubakar Abid, Alma Andersson, {\AA}ke Borg, Jonas Maaskola, Joakim Lundeberg, and James Zou.
\newblock {Integrating Spatial Gene Expression and Breast Tumour Morphology via Deep Learning}.
\newblock {\em Nature Biomedical Engineering}, 4(8):827--834, 2020.

\bibitem{Hu2026HistoPrism}
Susu Hu, Qinghe Zeng, Nithya Bhasker, Jakob~Nikolas Kather, and Stefanie Speidel.
\newblock {HistoPrism: Unlocking Functional Pathway Analysis from Pan-Cancer Histology via Gene Expression Prediction}.
\newblock In {\em International Conference on Learning Representations}, 2026.

\bibitem{huang2025scalable}
Tinglin Huang, Tianyu Liu, Mehrtash Babadi, Wengong Jin, and Rex Ying.
\newblock {Scalable Generation of Spatial Transcriptomics from Histology Images via Whole-Slide Flow Matching}.
\newblock In {\em International Conference on Machine Learning}, 2025.

\bibitem{huang2024rankbygene}
Wentao Huang, Meilong Xu, Xiaoling Hu, Shahira Abousamra, Aniruddha Ganguly, Saarthak Kapse, Alisa Yurovsky, Prateek Prasanna, Tahsin Kurc, Joel Saltz, Michael~L. Miller, and Chao Chen.
\newblock {RankByGene: Gene-Guided Histopathology Representation Learning Through Cross-Modal Ranking Consistency}.
\newblock {\em Transactions on Medical Imaging}, pages 1--1, 2026.

\bibitem{jarvelin2002cumulated}
Kalervo J{\"a}rvelin and Jaana Kek{\"a}l{\"a}inen.
\newblock {Cumulated Gain-based Evaluation of IR Techniques}.
\newblock In {\em ACM Transactions on Information Systems}, pages 422--446, 2002.

\bibitem{jaume2024hest}
Guillaume Jaume, Paul Doucet, Andrew Song, Ming~Yang Lu, Cristina Almagro~P{\'e}rez, Sophia Wagner, Anurag Vaidya, Richard Chen, Drew Williamson, Ahrong Kim, et~al.
\newblock {Hest-1k: A dataset for Spatial Transcriptomics and Histology Image Analysis}.
\newblock {\em Neural Information Processing Systems}, 37:53798--53833, 2024.

\bibitem{jaume2024modeling}
Guillaume Jaume, Anurag Vaidya, Richard~J Chen, Drew~FK Williamson, Paul~Pu Liang, and Faisal Mahmood.
\newblock {Modeling Dense Multimodal Interactions Between Biological Pathways and Histology for Survival Prediction}.
\newblock In {\em Computer Vision and Pattern Recognition}, pages 11579--11590, 2024.

\bibitem{jia2024thitogene}
Yuran Jia, Junliang Liu, Li~Chen, Tianyi Zhao, and Yadong Wang.
\newblock {THItoGene: A Deep Learning Method for Predicting Spatial Transcriptomics from Histological Images}.
\newblock {\em Briefings in Bioinformatics}, 25(1):bbad464, 2024.

\bibitem{kanehisa2021kegg}
Minoru Kanehisa, Miho Furumichi, Yoko Sato, Mayu Ishiguro-Watanabe, and Mao Tanabe.
\newblock {KEGG: Integrating Viruses and Cellular Organisms}.
\newblock {\em Nucleic Acids Research}, 49(D1):D545--D551, 2021.

\bibitem{kerseviciute2023apear}
Ieva Kerseviciute and Juozas Gordevicius.
\newblock {aPEAR: an R Package for Autonomous Visualization of Pathway Enrichment Networks}.
\newblock {\em Bioinformatics}, 39(11):btad672, 2023.

\bibitem{adam}
Diederik~P Kingma and Jimmy Ba.
\newblock {Adam: A Method for Stochastic Optimization}.
\newblock {\em ArXiv preprint arXiv:1412.6980}, 2014.

\bibitem{lu2024visual}
Ming~Y Lu, Bowen Chen, Drew~FK Williamson, Richard~J Chen, Ivy Liang, Tong Ding, Guillaume Jaume, Igor Odintsov, Long~Phi Le, Georg Gerber, et~al.
\newblock {A Visual-Language Foundation Model for Computational Pathology}.
\newblock {\em Nature medicine}, pages 863--874, 2024.

\bibitem{majumder2026pearl}
Sejuti Majumder, Saarthak Kapse, Moinak Bhattacharya, Xuan Xu, Alisa Yurovsky, and Prateek Prasanna.
\newblock {PEaRL: Pathway-Enhanced Representation Learning for Gene and Pathway Expression Prediction from Histology}.
\newblock In {\em Winter Conference on Applications of Computer Vision}, pages 8052--8062, 2026.

\bibitem{Maleki2019MethodCI}
Farhad Maleki, Katie~L. Ovens, Elham babazadeh Rezaei, Alan~M. Rosenberg, and Anthony~J. Kusalik.
\newblock {Method Choice in Gene Set Analysis Has Important Consequences for Analysis Outcome}.
\newblock In {\em Bioinformatics}, 2019.

\bibitem{mann1947test}
Henry~B Mann and Donald~R Whitney.
\newblock {On a Test of Whether One of Two Random Variables is Stochastically Larger Than the Other}.
\newblock {\em The Annals of Mathematical Statistics}, pages 50--60, 1947.

\bibitem{marx2021method}
Vivien Marx.
\newblock {Method of the Year: Spatially Resolved Transcriptomics}.
\newblock {\em Nature Methods}, 18(1):9--14, 2021.

\bibitem{nishimura2026cell}
Kazuya Nishimura, Ryoma Bise, Shinnosuke Matsuo, Haruka Hirose, and Yasuhiro Kojima.
\newblock {Cell-Type Prototype-informed Neural Network for Gene Expression Estimation from Pathology Images}.
\newblock {\em arXiv preprint arXiv:2603.18461}, 2026.

\bibitem{nishimura2025learning}
Kazuya Nishimura, Haruka Hirose, Ryoma Bise, Kaito Shiku, and Yasuhiro Kojima.
\newblock {Learning Relative Gene Expression Trends from Pathology Images in Spatial Transcriptomics}.
\newblock In {\em Neural Information Processing Systems}, 2025.

\bibitem{noh2026pathclast}
Minho Noh, Sungkyung Lee, Sunghyun Kim, and Sangsoo Lim.
\newblock {PathCLAST: Pathway-augmented Contrastive Learning with Attention for Interpretable Spatial Transcriptomics}.
\newblock {\em Briefings in Bioinformatics}, 27(1):bbag029, 2026.

\bibitem{pang2021leveraging}
Minxing Pang, Kenong Su, and Mingyao Li.
\newblock {Leveraging Information in Spatial Transcriptomics to Predict Super-resolution Gene Expression from Histology Images in Tumors}.
\newblock {\em BioRxiv}, pages 2021--11, 2021.

\bibitem{ritchie2015limma}
Matthew~E Ritchie, Belinda Phipson, DI~Wu, Yifang Hu, Charity~W Law, Wei Shi, and Gordon~K Smyth.
\newblock {Limma Powers Differential Expression Analyses for RNA-sequencing and Microarray Studies}.
\newblock {\em Nucleic Acids Research}, pages e47--e47, 2015.

\bibitem{robinson2010edger}
Mark~D Robinson, Davis~J McCarthy, and Gordon~K Smyth.
\newblock {edgeR: a Bioconductor Package for Differential Expression Analysis of Digital Gene Expression data}.
\newblock {\em Bioinformatics}, pages 139--140, 2010.

\bibitem{shiku2026auxiliary}
Kaito Shiku, Kazuya Nishimura, Shinnosuke Matsuo, Yasuhiro Kojima, and Ryoma Bise.
\newblock {Auxiliary Gene Learning: Spatial Gene Expression Estimation by Auxiliary Gene Selection}.
\newblock In {\em Association for the Advancement of Artificial Intelligence}, volume~40, pages 9015--9023, 2026.

\bibitem{stuart2019comprehensive}
Tim Stuart, Andrew Butler, Paul Hoffman, Christoph Hafemeister, Efthymia Papalexi, William~M Mauck, Yuhan Hao, Marlon Stoeckius, Peter Smibert, and Rahul Satija.
\newblock {Comprehensive Integration of Single-cell Data}.
\newblock {\em cell}, 177(7):1888--1902, 2019.

\bibitem{walker2022deciphering}
Benjamin~L Walker, Zixuan Cang, Honglei Ren, Eric Bourgain-Chang, and Qing Nie.
\newblock {Deciphering Tissue Structure and Function Using Spatial Transcriptomics}.
\newblock {\em Communications Biology}, 5(1):220, 2022.

\bibitem{Wang2024M2ORTMR}
Hongyi Wang, Xiuju Du, Jing Liu, Shuyi Ouyang, Yen wei Chen, and Lanfen Lin.
\newblock {M2ORT: Many-To-One Regression Transformer for Spatial Transcriptomics Prediction from Histopathology Images}.
\newblock {\em Association for the Advancement of Artificial Intelligence}, 2025.

\bibitem{williams2022introduction}
Cameron~G Williams, Hyun~Jae Lee, Takahiro Asatsuma, Roser Vento-Tormo, and Ashraful Haque.
\newblock {An Introduction to Spatial Transcriptomics for Biomedical Research}.
\newblock {\em Genome Medicine}, 14(1):68, 2022.

\bibitem{xie2023spatially}
Ronald Xie, Kuan Pang, Sai Chung, Catia Perciani, Sonya MacParland, Bo~Wang, and Gary Bader.
\newblock {Spatially Resolved Gene Expression Prediction from Histology Images via Bi-modal Contrastive Learning}.
\newblock {\em Neural Information Processing Systems}, 36:70626--70637, 2023.

\bibitem{xu2024whole}
Hanwen Xu, Naoto Usuyama, Jaspreet Bagga, Sheng Zhang, Rajesh Rao, Tristan Naumann, Cliff Wong, Zelalem Gero, Javier Gonz{\'a}lez, Yu~Gu, et~al.
\newblock {A Whole-slide Foundation Model for Digital Pathology from Real-world Data}.
\newblock {\em Nature}, 630(8015):181--188, 2024.

\bibitem{yang2024spatial}
Yan Yang, Md~Zakir Hossain, Eric Stone, and Shafin Rahman.
\newblock {Spatial Transcriptomics Analysis of Gene Expression Prediction Using Exemplar Guided Graph Neural Network}.
\newblock {\em Pattern Recognition}, 145:109966, 2024.

\bibitem{yang2023exemplar}
Yan Yang, Md~Zakir Hossain, Eric~A Stone, and Shafin Rahman.
\newblock {Exemplar Guided Deep Neural Network for Spatial Transcriptomics Analysis of Gene Expression Prediction}.
\newblock In {\em Winter Conference on Applications of Computer Vision}, pages 5039--5048, 2023.

\bibitem{zeng2022spatial}
Yuansong Zeng, Zhuoyi Wei, Weijiang Yu, Rui Yin, Yuchen Yuan, Bingling Li, Zhonghui Tang, Yutong Lu, and Yuedong Yang.
\newblock {Spatial Transcriptomics Prediction from Histology Jointly Through Transformer and Graph Neural Networks}.
\newblock {\em Briefings in Bioinformatics}, 23(5):bbac297, 2022.

\bibitem{zhu2025diffusion}
Sichen Zhu, Yuchen Zhu, Molei Tao, and Peng Qiu.
\newblock {Diffusion Generative Modeling for Spatially Resolved Gene Expression Inference from Histology Images}.
\newblock In {\em International Conference on Learning Representations}, 2025.

\bibitem{zimmermann2024virchow2}
Eric Zimmermann, Eugene Vorontsov, Julian Viret, Adam Casson, Michal Zelechowski, George Shaikovski, Neil Tenenholtz, James Hall, Thomas Fuchs, Nicolo Fusi, Siqi Liu, and Kristen Severson.
\newblock {Virchow2: Scaling Self-Supervised Mixed Magnification Models in Pathology}.
\newblock {\em arXiv preprint arXiv:2408.00738}, 2024.

\end{thebibliography}

\clearpage

\appendix





\section{Supplementary Materials}
\subsection{Limitations and Future Work}
Our results demonstrate the effectiveness of explicitly optimizing DEG rankings compared with conventional objectives that do not consider gene rankings.
However, our current objective optimizes the overall ranking agreement across all genes.
Developing ranking objectives that place greater emphasis on top-ranked DEGs, which are particularly important for downstream biological analysis, is a promising direction for future work.

Additionally, the absolute DEG-ranking performance on pathologist-annotated regions remains relatively low.
This may partly reflect the inherent difficulty of histology-based gene expression prediction on challenging datasets such as Her2st, indicating room for further improvement.

\subsection{Evaluation and Biological Interpretation of Significantly Enriched Pathways}
In the main paper, we conducted a Pathway Enrichment Overlap Analysis by calculating the Jaccard index over the top 200 enriched pathways.
We further evaluated pathway enrichment using only significantly enriched pathways on the Her2st dataset.
Specifically, we restricted the analysis to pathways with FDR-adjusted $p < 0.05$ and calculated the Jaccard index.
Furthermore, we ranked pathways according to their enrichment combined scores and used SCC to quantify the agreement between the ground-truth and predicted rankings.
We also report PCC to evaluate the agreement between the ground-truth and predicted combined scores.

As shown in Table~\ref{tab:pathway_appendix}, IDER achieves the best performance across all three metrics. In particular, the improvement in SCC demonstrates that IDER better preserves the ranking of enriched pathways, highlighting its effectiveness in capturing pathway-level ranking agreement beyond pathway-set overlap.

The recovered pathways are also consistent with known biology.
For example, invasive cancer regions exhibit highly ranked ERBB2 signaling pathways associated with HER2-positive breast cancer, while immune infiltrate regions exhibit highly ranked tumor-immune pathways related to antigen presentation, T-cell activation, and PD-1 signaling.
These findings suggest that the improved pathway-ranking agreement reflects biologically relevant signals rather than merely increased overlap of enriched pathways.

\begin{table}[!htbp]
\centering
\caption{\textbf{Comparison of pathway enrichment performance across different training objectives on the Her2st dataset.} Evaluation is restricted to significantly enriched pathways (FDR-adjusted $p < 0.05$). Jaccard measures the overlap of significant pathway sets, while PCC and SCC measure the agreement of enrichment combined scores and pathway rankings, respectively, between ground-truth and predicted gene expression.}
\label{tab:pathway_appendix}
\scalebox{0.85}{
\begin{tabular}{c|ccc}
\toprule
Objective & Jaccard & PCC & SCC \\
\midrule
MSE       & 0.071 & 0.157 & 0.010 \\
PCC       & 0.058 & 0.161 & 0.020 \\
MSE \& PCC & 0.062 & 0.222 & -0.139 \\
Poisson   & 0.073 & 0.374 & 0.088 \\
NB        & 0.064 & 0.196 & -0.045 \\
STRank    & 0.086 & 0.169 & -0.025 \\
\rowcolor{gray!15}
\textbf{Ours} & \textbf{0.227} & \textbf{0.480} & \textbf{0.271} \\
\bottomrule
\end{tabular}
}
\end{table}

\FloatBarrier

\subsection{Generalization Across Multiple Pathology Foundation Models}
To evaluate whether our method generalizes beyond the CONCH-specific morphological representations used for group construction during training, we conducted additional evaluations using tissue morphology groups constructed with different pathology foundation models.
Specifically, we evaluated our method on the Ovary dataset using four state-of-the-art pathology foundation models: UNIv2~\cite{chen2024towards}, Virchow2~\cite{zimmermann2024virchow2}, OmniCLIP~\cite{chen2025visual}, and GigaPath~\cite{xu2024whole}.

As shown in Table~\ref{tab:foundation_model_comparison}, our method consistently achieves the best DEG-ranking performance across morphology groups constructed using all four foundation models.
These results demonstrate that IDER, although trained using CONCH-derived morphology groups, is not strongly dependent on CONCH-specific group definitions and generalizes to morphological contrasts captured by different pathology foundation models.

\begin{table}[!htbp]
    \centering
    \caption{\textbf{Comparison of performance across different pathology foundation models} against methods optimized with conventional objective functions.}
    \scalebox{0.85}{
    \begin{tabular}{c|cccc}
        \toprule
        Objective
        & UNIv2
        & Virchow2
        & OmniCLIP
        & GigaPath \\
        \midrule
        MSE
            & 0.595 & 0.550 & 0.607 & 0.598 \\
        PCC
            & 0.593 & 0.551 & 0.604 & 0.592 \\
        MSE $\&$ PCC
            & 0.612 & 0.569 & 0.627 & 0.614 \\
        Poisson
            & 0.506 & 0.471 & 0.510 & 0.473 \\
        NB
            & 0.323 & 0.288 & 0.302 & 0.281 \\
        STRank
            & 0.502 & 0.470 & 0.503 & 0.468 \\
        \rowcolor{gray!15}
        \textbf{Ours}
            & \textbf{0.674}
            & \textbf{0.633}
            & \textbf{0.692}
            & \textbf{0.672} \\
        \bottomrule
    \end{tabular}
    }
    \label{tab:foundation_model_comparison}
    \vspace{-4mm}
\end{table}

\FloatBarrier

\subsection{DEG Ranking Performance on Highly Variable Genes}
IDER allows DEG-ranking performance to be evaluated on any subset of genes at inference time.
To assess its effectiveness on highly variable genes (HVGs), we additionally evaluated IDER on the Her2st dataset using the top 2,000 and 5,000 HVGs, which are commonly used in transcriptomic analyses.

As shown in Table~\ref{tab:hvg_comparison}, IDER consistently achieves the best performance in terms of both SCC and nDCG@200 for the top 2,000 and 5,000 HVGs.
These results demonstrate that IDER remains effective when focusing on HVGs, indicating that its improvements are not solely attributable to the inclusion of all genes.

\begin{table}[!htbp]
    \centering
    \caption{\textbf{Comparison of DEG ranking performance across different numbers of highly variable genes (HVGs)} against methods optimized with conventional objective functions, in terms of SCC$^{\scriptsize \mathrm{DEG}}$ and $\mathrm{nDCG}^{\mathrm{DEG}}$@200. Due to space limitations, $\mathrm{nDCG}^{\mathrm{DEG}}$@200 is abbreviated as $\mathrm{nDCG}^{\mathrm{DEG}}$.}
    \scalebox{0.85}{
    \begin{tabular}{c|cc|cc}
        \toprule
        \multirow{2}{*}{Objective}
        & \multicolumn{2}{c|}{\textbf{HVG2000}}
        & \multicolumn{2}{c}{\textbf{HVG5000}} \\
        \cmidrule(lr){2-3} \cmidrule(lr){4-5}
        & SCC$^{\scriptsize \mathrm{DEG}}$ & $\mathrm{nDCG}^{\mathrm{DEG}}$
        & SCC$^{\scriptsize \mathrm{DEG}}$ & $\mathrm{nDCG}^{\mathrm{DEG}}$ \\
        \midrule
        MSE
            & 0.205 & 0.231 & 0.221 & 0.145 \\
        PCC
            & 0.168 & 0.203 & 0.178 & 0.126 \\
        MSE $\&$ PCC
            & 0.193 & 0.212 & 0.188 & 0.123 \\
        Poisson
            & 0.149 & 0.215 & 0.143 & 0.145 \\
        NB
            & 0.177 & 0.216 & 0.173 & 0.153 \\
        STRank
            & 0.161 & 0.217 & 0.143 & 0.154 \\
        \rowcolor{gray!15}
        \textbf{Ours}
            & \textbf{0.351} & \textbf{0.343}
            & \textbf{0.362} & \textbf{0.286} \\
        \bottomrule
    \end{tabular}
    }
    \label{tab:hvg_comparison}
    \vspace{-4mm}
\end{table}

\FloatBarrier

\subsection{Comparative experiments reporting mean and standard deviation}
Due to space limitations, only the mean values from cross-validation were reported in the main text for the comparative experiments under the cluster-based DEG setting and the real annotation-based DEG setting. In this section, we additionally present the results with standard deviations, where the results for the cluster-based DEG setting are shown in Table~\ref{tab:combined_cluster} and those for the real annotation-based DEG setting are shown in Table~\ref{tab:combined_tissue}.
Furthermore, for the real annotation-based DEG setting, we also report the $\mathrm{nDCG}^{\mathrm{DEG}}$@50 and @100 metrics, which were omitted in the main text.

\begin{table}[!htbp]
\centering
\caption{\textbf{Comparison of DEG ranking performance in the cluster-based DEG setting.}}
\label{tab:combined_cluster}
\scalebox{0.6}{
\begin{tabular}{c|c|ccc}
\toprule
\multirow{2}{*}{Objective} 
& \multirow{2}{*}{SCC$^{\scriptsize \mathrm{DEG}}$} 
& \multicolumn{3}{c}{$\mathrm{nDCG}^{\mathrm{DEG}}$} \\
\cmidrule(lr){3-5}
& & @50 & @100 & @200 \\
\midrule

\multicolumn{5}{c}{\textbf{Ovary}} \\
\midrule
MSE        & 0.587$\pm$0.009 & 0.415$\pm$0.003 & 0.446$\pm$0.004 & 0.434$\pm$0.005 \\
PCC        & 0.584$\pm$0.011 & 0.425$\pm$0.006 & 0.451$\pm$0.003 & 0.451$\pm$0.006 \\
MSE \& PCC & 0.609$\pm$0.014 & 0.437$\pm$0.011 & 0.465$\pm$0.010 & 0.453$\pm$0.006 \\
Poisson    & 0.460$\pm$0.009 & 0.363$\pm$0.004 & 0.401$\pm$0.004 & 0.422$\pm$0.008 \\
NB         & 0.245$\pm$0.016 & 0.404$\pm$0.012 & 0.415$\pm$0.012 & 0.417$\pm$0.009 \\
STRank    & 0.457$\pm$0.018 & 0.359$\pm$0.010 & 0.399$\pm$0.006 & 0.416$\pm$0.005 \\
\rowcolor{gray!15}
\textbf{Ours} & \textbf{0.686}$\pm$0.004 & \textbf{0.486}$\pm$0.026 & \textbf{0.515}$\pm$0.011 & \textbf{0.531}$\pm$0.011 \\

\midrule
\multicolumn{5}{c}{\textbf{Lymph Node}} \\
\midrule
MSE        & 0.624$\pm$0.013 & 0.407$\pm$0.008 & 0.434$\pm$0.008 & 0.477$\pm$0.006 \\
PCC        & 0.629$\pm$0.007 & 0.439$\pm$0.012 & 0.478$\pm$0.010 & 0.521$\pm$0.007 \\
MSE \& PCC & 0.650$\pm$0.002 & 0.477$\pm$0.010 & 0.511$\pm$0.005 & 0.548$\pm$0.002 \\
Poisson    & 0.234$\pm$0.012 & 0.448$\pm$0.020 & 0.493$\pm$0.014 & 0.501$\pm$0.009 \\
NB         & 0.090$\pm$0.018 & 0.541$\pm$0.004 & 0.548$\pm$0.006 & 0.534$\pm$0.004 \\
STRank    & 0.300$\pm$0.009 & 0.375$\pm$0.014 & 0.425$\pm$0.003 & 0.446$\pm$0.008 \\
\rowcolor{gray!15}
\textbf{Ours} & \textbf{0.676}$\pm$0.006 & \textbf{0.601}$\pm$0.011 & \textbf{0.641}$\pm$0.007 & \textbf{0.643}$\pm$0.007 \\

\midrule
\multicolumn{5}{c}{\textbf{Bowel}} \\
\midrule
MSE        & 0.709$\pm$0.007 & 0.364$\pm$0.005 & 0.432$\pm$0.006 & 0.468$\pm$0.007 \\
PCC        & 0.710$\pm$0.001 & 0.384$\pm$0.005 & 0.461$\pm$0.004 & 0.494$\pm$0.006 \\
MSE \& PCC & 0.714$\pm$0.001 & 0.383$\pm$0.008 & 0.459$\pm$0.005 & 0.496$\pm$0.004 \\
Poisson    & 0.694$\pm$0.001 & 0.440$\pm$0.009 & 0.476$\pm$0.006 & 0.492$\pm$0.007 \\
NB         & 0.601$\pm$0.003 & \textbf{0.494}$\pm$0.019 & 0.533$\pm$0.013 & 0.546$\pm$0.008 \\
STRank    & 0.687$\pm$0.002 & 0.435$\pm$0.011 & 0.470$\pm$0.008 & 0.498$\pm$0.001 \\
\rowcolor{gray!15}
\textbf{Ours} & \textbf{0.731}$\pm$0.005 & 0.481$\pm$0.017 & \textbf{0.565}$\pm$0.012 & \textbf{0.595}$\pm$0.012 \\

\bottomrule
\end{tabular}
}
\vspace{-4mm}
\end{table}

\begin{table}[!htbp]
\centering
\caption{\textbf{Comparison of DEG ranking performance in the real annotation-based DEG setting}.}
\label{tab:combined_tissue}
\scalebox{0.6}{
\begin{tabular}{c|c|ccc}
\toprule
\multirow{2}{*}{Objective} 
& \multirow{2}{*}{SCC$^{\scriptsize \mathrm{DEG}}$} 
& \multicolumn{3}{c}{$\mathrm{nDCG}^{\mathrm{DEG}}$} \\
\cmidrule(lr){3-5}
& & @50 & @100 & @200 \\
\midrule

\multicolumn{5}{c}{\textbf{Average}} \\
\midrule
MSE          & 0.239$\pm$0.085 & 0.053$\pm$0.050 & 0.065$\pm$0.054 & 0.084$\pm$0.053 \\
PCC          & 0.181$\pm$0.119 & 0.060$\pm$0.057 & 0.062$\pm$0.055 & 0.073$\pm$0.047 \\
MSE \& PCC   & 0.195$\pm$0.117 & 0.051$\pm$0.049 & 0.061$\pm$0.054 & 0.073$\pm$0.051 \\
Poisson      & 0.116$\pm$0.079 & 0.086$\pm$0.060 & 0.095$\pm$0.060 & 0.101$\pm$0.057 \\
NB           & 0.137$\pm$0.138 & 0.072$\pm$0.064 & 0.086$\pm$0.070 & 0.098$\pm$0.066 \\
STRank      & 0.101$\pm$0.104 & 0.085$\pm$0.081 & 0.100$\pm$0.086 & 0.109$\pm$0.076 \\
\rowcolor{gray!15}
\textbf{Ours} & \textbf{0.353}$\pm$0.084 & \textbf{0.212}$\pm$0.104 & \textbf{0.230}$\pm$0.101 & \textbf{0.247}$\pm$0.067 \\

\midrule
\multicolumn{5}{c}{\textbf{Adipose tissue}} \\
\midrule
MSE          & 0.393$\pm$0.080 & 0.083$\pm$0.083 & 0.068$\pm$0.052 & 0.069$\pm$0.034 \\
PCC          & 0.297$\pm$0.177 & 0.071$\pm$0.071 & 0.057$\pm$0.053 & 0.060$\pm$0.034 \\
MSE \& PCC   & 0.317$\pm$0.102 & 0.083$\pm$0.083 & 0.069$\pm$0.057 & 0.076$\pm$0.043 \\
Poisson      & 0.123$\pm$0.052 & 0.044$\pm$0.035 & 0.053$\pm$0.026 & 0.057$\pm$0.014 \\
NB           & 0.145$\pm$0.196 & 0.025$\pm$0.025 & 0.036$\pm$0.027 & 0.038$\pm$0.018 \\
STRank      & 0.053$\pm$0.006 & 0.081$\pm$0.020 & 0.085$\pm$0.001 & 0.083$\pm$0.014 \\
\rowcolor{gray!15}
\textbf{Ours} & \textbf{0.475}$\pm$0.099 & \textbf{0.092}$\pm$0.029 & \textbf{0.087}$\pm$0.021 & \textbf{0.100}$\pm$0.031 \\

\midrule
\multicolumn{5}{c}{\textbf{Connective tissue}} \\
\midrule
MSE          & 0.206$\pm$0.054 & 0.053$\pm$0.039 & 0.063$\pm$0.042 & 0.080$\pm$0.041 \\
PCC          & 0.143$\pm$0.107 & 0.052$\pm$0.058 & 0.055$\pm$0.063 & 0.066$\pm$0.045 \\
MSE \& PCC   & 0.208$\pm$0.120 & 0.027$\pm$0.039 & 0.039$\pm$0.038 & 0.048$\pm$0.028 \\
Poisson      & 0.105$\pm$0.092 & 0.080$\pm$0.076 & 0.099$\pm$0.079 & 0.109$\pm$0.072 \\
NB           & 0.127$\pm$0.133 & 0.062$\pm$0.076 & 0.079$\pm$0.093 & 0.097$\pm$0.083 \\
STRank      & 0.114$\pm$0.119 & 0.102$\pm$0.151 & 0.107$\pm$0.137 & 0.121$\pm$0.112 \\
\rowcolor{gray!15}
\textbf{Ours} & \textbf{0.313}$\pm$0.133 & \textbf{0.210}$\pm$0.185 & \textbf{0.189}$\pm$0.166 & \textbf{0.164}$\pm$0.123 \\

\midrule
\multicolumn{5}{c}{\textbf{Immune infiltrate}} \\
\midrule
MSE          & 0.197$\pm$0.039 & 0.080$\pm$0.033 & 0.130$\pm$0.037 & 0.176$\pm$0.021 \\
PCC          & 0.132$\pm$0.040 & 0.152$\pm$0.124 & 0.165$\pm$0.096 & 0.199$\pm$0.074 \\
MSE \& PCC   & 0.134$\pm$0.112 & 0.167$\pm$0.001 & 0.214$\pm$0.030 & 0.239$\pm$0.014 \\
Poisson      & 0.191$\pm$0.025 & 0.191$\pm$0.059 & 0.203$\pm$0.060 & 0.228$\pm$0.074 \\
NB           & 0.190$\pm$0.010 & 0.219$\pm$0.078 & 0.256$\pm$0.113 & 0.269$\pm$0.089 \\
STRank      & 0.189$\pm$0.001 & 0.191$\pm$0.070 & 0.223$\pm$0.043 & 0.227$\pm$0.043 \\
\rowcolor{gray!15}
\textbf{Ours} & \textbf{0.263}$\pm$0.006 & \textbf{0.387}$\pm$0.002 & \textbf{0.426}$\pm$0.026 & \textbf{0.446}$\pm$0.026 \\

\midrule
\multicolumn{5}{c}{\textbf{Invasive cancer}} \\
\midrule
MSE          & 0.272$\pm$0.104 & 0.056$\pm$0.060 & 0.065$\pm$0.076 & 0.082$\pm$0.082 \\
PCC          & 0.182$\pm$0.123 & 0.029$\pm$0.026 & 0.030$\pm$0.027 & 0.035$\pm$0.030 \\
MSE \& PCC   & 0.216$\pm$0.125 & 0.025$\pm$0.029 & 0.031$\pm$0.034 & 0.043$\pm$0.042 \\
Poisson      & 0.114$\pm$0.097 & 0.083$\pm$0.062 & 0.081$\pm$0.051 & 0.081$\pm$0.040 \\
NB           & 0.120$\pm$0.185 & 0.057$\pm$0.047 & 0.062$\pm$0.045 & 0.073$\pm$0.041 \\
STRank      & 0.096$\pm$0.127 & 0.040$\pm$0.064 & 0.065$\pm$0.070 & 0.077$\pm$0.063 \\
\rowcolor{gray!15}
\textbf{Ours} & \textbf{0.357}$\pm$0.125 & \textbf{0.188}$\pm$0.071 & \textbf{0.245}$\pm$0.062 & \textbf{0.301}$\pm$0.068 \\

\bottomrule
\end{tabular}
}
\vspace{-4mm}
\end{table}

\FloatBarrier

\subsection{Ablation Results for All Evaluation Metrics}
Due to space limitations, only a subset of evaluation metrics for the ablation experiments was reported in the main text. Here, we additionally report the performance on the omitted metrics, as shown in Table~\ref{tab:comparison_ablation}.

\begin{table}[!htbp]
\centering
\caption{\textbf{Ablation experiments across datasets.}}
\label{tab:comparison_ablation}
\scalebox{0.8}{
\begin{tabular}{c|cc|c|ccc}
\toprule
\multirow{2}{*}{Objective}
& \multicolumn{2}{c|}{Setting}
& \multirow{2}{*}{SCC$^{\scriptsize \mathrm{DEG}}$}
& \multicolumn{3}{c}{$\mathrm{nDCG}^{\mathrm{DEG}}$} \\
\cmidrule(lr){2-3} \cmidrule(lr){5-7}
& U stat. & DEG rank. & & @50 & @100 & @200 \\
\midrule

\multicolumn{7}{c}{\textbf{Ovary}} \\
\midrule
Baseline & \ding{55} & \ding{55} & 0.584$\pm$0.011 & 0.425$\pm$0.006 & 0.451$\pm$0.003 & 0.451$\pm$0.006\\
Ours w/o DEG rank. & \ding{51} & \ding{55} & 0.587$\pm$0.005 & 0.439$\pm$0.009 & 0.459$\pm$0.004 & 0.454$\pm$0.001 \\
\rowcolor{gray!15}
\textbf{Ours} & \ding{51} & \ding{51} & \textbf{0.686}$\pm$0.004 & \textbf{0.486}$\pm$0.026 & \textbf{0.515}$\pm$0.011 & \textbf{0.531}$\pm$0.011 \\

\midrule
\multicolumn{7}{c}{\textbf{Lymph Node}} \\
\midrule
Baseline & \ding{55} & \ding{55} & 0.629$\pm$0.007 & 0.439$\pm$0.012 & 0.478$\pm$0.010 & 0.521$\pm$0.007 \\
Ours w/o DEG rank. & \ding{51} & \ding{55} & 0.617$\pm$0.004 & 0.440$\pm$0.009 & 0.475$\pm$0.004 & 0.521$\pm$0.003 \\
\rowcolor{gray!15}
\textbf{Ours} & \ding{51} & \ding{51} & \textbf{0.676}$\pm$0.006 & \textbf{0.601}$\pm$0.011 & \textbf{0.641}$\pm$0.007 & \textbf{0.643}$\pm$0.007 \\

\midrule
\multicolumn{7}{c}{\textbf{Bowel}} \\
\midrule
Baseline & \ding{55} & \ding{55} & 0.710$\pm$0.001 & 0.384$\pm$0.005 & 0.461$\pm$0.004 & 0.494$\pm$0.006\\
Ours w/o DEG rank.& \ding{51} & \ding{55} & 0.710$\pm$0.004 & 0.379$\pm$0.013 & 0.464$\pm$0.015 & 0.491$\pm$0.010 \\
\rowcolor{gray!15}
\textbf{Ours} & \ding{51} & \ding{51} & \textbf{0.731}$\pm$0.005 & \textbf{0.481}$\pm$0.017 & \textbf{0.565}$\pm$0.012 & \textbf{0.595}$\pm$0.012 \\

\midrule
\multicolumn{7}{c}{\textbf{Her2st}} \\
\midrule
Baseline & \ding{55} & \ding{55} & 0.181$\pm$0.119 & 0.060$\pm$0.057 & 0.062$\pm$0.055 & 0.073$\pm$0.047 \\
Ours w/o DEG rank. & \ding{51} & \ding{55} & 0.178$\pm$0.113 & 0.068$\pm$0.060 & 0.066$\pm$0.051 & 0.077$\pm$0.046 \\
\rowcolor{gray!15}
\textbf{Ours} & \ding{51} & \ding{51} & \textbf{0.353}$\pm$0.084 & \textbf{0.212}$\pm$0.104 & \textbf{0.230}$\pm$0.101 & \textbf{0.247}$\pm$0.067 \\

\bottomrule
\end{tabular}
}
\vspace{-2mm}
\end{table}

\FloatBarrier

\subsection{Processing Details of Dataset}
\label{sec:preprocessing}
\noindent
{\bf Bowel, Ovary, and Lymph Node Datasets.}
We removed spots with fewer than 200 detected genes and genes detected in fewer than 3 spots.
We then applied spot-wise total-count normalization followed by a log1p transformation.
No imputation or spatial smoothing was performed.
After preprocessing, 18,066, 18,054, and 17,564 genes remained for Bowel, Ovary, and Lymph Node, respectively.
Each dataset was split into training, validation, and test sets at a ratio of 3:1:1.

\noindent
{\bf Her2st Dataset.}
We first retained genes shared across all patients.
We then removed spots with fewer than 200 detected genes and genes detected in fewer than 3 spots.
We applied a log1p transformation.
No imputation or spatial smoothing was performed.
After preprocessing, 11,871 genes remained.
We performed four-fold patient-wise cross-validation on patients A--H, splitting them into training, validation, and test sets at a ratio of 2:1:1.

\subsection{Computer Resources}
All experiments were conducted on a desktop computer with the following specifications:

\begin{itemize}
    \item CPU: Intel(R) Xeon(R) Gold 5122 CPU (8 cores, 16 threads)
    \item GPU: NVIDIA RTX A6000 (48 GB GPU memory)
    \item Memory: ~128 GB
\end{itemize}

\subsection{Licenses for Existing Assets}
We implemented our method using PyTorch~\cite{Paszke2019PyTorchAI}, which is released under the BSD 3-Clause License.
For the experiments, we used publicly available datasets, including HEST-1k (CC BY-NC-SA 4.0) and Her2st (CC BY 4.0). 

\FloatBarrier



\end{document}